# Semantic Network Model for Sign Language Comprehension

Xinchen Kang (22f3d431-dcc0-42a5-8e2b-c26464e0654d)

*Beijing Union University, China*

Dengfeng Yao (ae88317c-d091-4f41-97f1-b1e6be00ca68)

*Beijing Union University, China*

Minghu Jiang (ea1cc43b-eee9-4185-8d97-edeac9186268)

*Tsinghua University, China*

Yunlong Huang (cc1ddbf2-64b9-4c51-b553-0ebe52a8d645)

*Tsinghua University, China*

Fanshu Li (64642396-6e95-456f-a4f2-47ff81a23d6e)

*Beijing Union University, China*

**ABSTRACT**

*In this study, the authors propose a computational cognitive model for sign language (SL) perception and comprehension with detailed algorithmic descriptions based on cognitive functionalities in human language processing. The semantic network model (SNM) that represents semantic relations between concepts is used as a form of knowledge representation. The proposed model is applied in the comprehension of sign language for classifier predicates. The spreading activation search method is initiated by labeling a set of source nodes (e.g. concepts in the semantic network) with weights or "activation," and then iteratively propagating or "spreading" that activation out to other nodes linked to the source nodes. The results demonstrate that the proposed search method improves the performance of sign language comprehension in the SNM.*

**INTRODUCTION**

Sign language (SL) comprehension is a fundamental task for computational linguists. Two types of algorithms have been proposed: (1) rule-based methods (Supalla, 1982), and (2) statistical methods (Bauer & Heinz, 2000; Huenerfauth, 2005). Rule-based methods lack the capability of planning the elements in the entire scene (Liddell, 2003). The method of modeling infinite natural language input through finite rules, especially minor rules, barely meets all requirements of SL processing (Yao et al., 2017). Therefore, statistical methods are the preferred type of algorithm for SL comprehension. Statistical models can be applied to spoken languages. Given the abundant data resources of spoken languages in the digitalized Internet age, statistical models can be applied readily. However, the raw and annotated corpora of SLs are insufficient because collecting and annotating SL videos are tedious and difficult. Data sparsity consequently remains as the most serious problem when applying statistical models onto SLs. For example, the real-time factor (RTF) of the SL video corpus is 100; that is, an hour corpus requires at least 100 hours of annotation (Dreuw et al., 2008b).

Simulating SL comprehension using traditional statistical models and machine-learning methods isdifficult. Thus, reliable methods for establishing a signer's 3-D model (which is the process of developing a mathematical representation of any three-dimensional surface of moving trajectories of signers in the space for SL via specialized software) for SL corpus building and technologies for annotating a large-scale SL video corpus automatically must be developed. Unlike the spoken language that is "a set of values that change with the passage of time" (Huenerfauth, 2005), SL does not have a writing system and thus cannot be saved in any form of written texts.

The natural language-processing system relies on texts to process spoken languages. This system records only the written text that corresponds to speech flows and relies only on the literacy of the user. On the other hand, the SL system comprises information from multiple modalities. Examples of such information are the hand shape, hand location, hand movement, hand orientation, head tilting, shoulder tilting, eye gazing, body gestures, and facial expressions. The considerable information from multiple channels in SL conveys linguistic meaning. This multi-modality nature of SL poses difficulties for the coding of SLs into a linear single-channeled character string. In addition, SLs have writing systems, such as the *Sign Writing* system (Sutton, 2010), *ASL-phabet* (Supalla et al., 2008), and *HamNoSys* (Prillwitz et al., 1989). However, these systems have a limited number of users (Johnston, 2004).

Many linguistic details are lost because of the multi-modality nature of SL during the translation of SL into its corresponding writing system. SLs may be understood by directly matching the visual–spatial characteristics of SL with the semantic units in the brain rather than applying written texts as an interpreting medium. Here, semantic units are generally used for processing natural languages; these units or nodes contain some information, which are used as knowledge representations form semantic units (Geva et al., 2000). Such direct matching also represents the most natural way of comprehending SLs in the brain. From this perspective, the authors present a computational cognitive model for SL comprehension that is based on the cognitive functionalities of the human brain combined with a knowledge representation theory of artificial intelligence (Shuklin, 2001).

Visual–spatial mechanisms are exploited to express the grammatical structures and functions in SL. Visual–spatial perception, memory, and mental transformations are prerequisites to grammatical processing in SL (Emmorey & Corina, 1990) and are central to visual mental imagery (Farah, 1988). A series of experiments have been conducted to investigate visual attention (Neville et al., 1998). Movement recognition in peripheral vision is important in sign perception because the signers mainly look at the face instead of tracking the hands when they communicate through SL (Siple, 1978). Therefore, identification of lexical-level information depends on the peripheral vision system when signs are produced. The recognition of movement directions is the selective function of peripheral vision (Bonnet, 1977).

Whether deaf people only have a strong peripheral vision or efficiently allocate attention to peripheral vision remains unclear. Stivalet et al. (1998) showed that visual attention processing can be changed by auditory deprivation. They determined that deaf people do not shift their attention when processing the information (i.e., alphabet set) presented in the central vision field, whereas hearing subjects must shift their attention to search for the alphabet set continuously. Smith et al. (1998) also found that lack of auditory input causes weak and selective (or highly distributed) visual attention among deaf children. Stivalet et al. (1998) proposed that effective visual processing is caused by intermodal sensory compensation; that is, the strong allocation of visual attention can be attributed to neuron reorganization caused by auditory deprivation from birth. Recent magnetic resonance imaging evidence supports this hypothesis (Bavelier et al., 2000).

These findings are selective attention cases, in which attention selectively processes certain stimuli but ignores other stimuli. The cases refer to the selective orientation and concentration of the senses (i.e., visual, auditory, taste, and tactile senses) and consciousness (i.e., awareness) of people on certain targets (towards other factors). Studies on attention have failed to describe human attention at the biological level in detail, as a person cannot focus continuously because the brain automatically suppresses activity when attention reaches its limits.

Emmorey and Reilly (2013) determined that when locations in a signing space (SL expressions streaks the space) function topographically, spatial changes tend to be noticed easily. Thus, location information indicating the spatial position of associated referents can be encoded and stored semantically in memory. However, spatial locations with a primary distinguishing function of referents are encoded in a different way and tend to be discarded from memory once the referential function is no longer required by context (Emmorey & Reilly, 2013). Bavelier et al. (2001) claimed that only the posterior middle temporal gyrus and the medial superior temporal cortex of deaf signers are highly active while perceiving movements in peripheral vision. This phenomenon is unobservable in hearing signers who have skillfully grasped signs, indicating that auditory deprivation results in a shift to stronger movement attention in the visual periphery. Deaf people can easily reply to the attention and visual monitoring of their peri-personal space (Bavelier et al., 2001).

Neville et al. (1998) determined that the classic language area in the left hemisphere, particularly the left perisylvian, of both deaf and hearing subjects is activated when reading English sentences. The right hemisphere, including the right perisylvian, of deaf people is also activated. They argued that spatial processing is of great importance to sign grammar. Thus, the SL comprehension process of deaf people employs neurons at both high and low levels in the neural network, which are connected with each other by edges, and generates high-level features via feature combination processes that are realized by combining the weight on the edges. For example, low-level visual edge features are assembled, processed, and sent to the high level to form the angle, shape, and other higher features (Bertasius et al., 2015). High-level neurons form features that gradually approximate the semantics in turn, such as simple shapes, simple targets, and real objects. The activation of high-level neurons during the reconstruction process also reacts with the low-level neurons and adjusts and corrects deviations and losses (Bertasius et al., 2015); a temporal pattern appears in the horizontal structure connection. The neurons can make predictions of the state at the next point of every time point through a horizontal connection based on the information of their current status (Hawkins et al., 2009).

## SEMANTIC NETWORK MODEL (SNM)

Model of semantic networks (SNs) are generally used for processing natural languages (Shuklin, 2001). SNs, as knowledge representations, are extensible and have been used to model mental disturbances (Geva et al., 2000). The semantic network (which is a network that represents semantic relations between concepts, is used as a form of knowledge representation, here it is based on SL information processing of human brain cortex. The edges connect different nodes in the network and represent the strength or weakness of the correlation. After being set up, the semantic network is stored in long-term memory for future retrieval and extraction to be encoded as semantic memory. Outside stimuli at a certain time can be the demand of a person on specific knowledge and information to activate the demand on the extraction of useful information of long-term memory (Sedikides & Skowronski, 1991). The activation process of the stored network works in a form of spreading in the memory (Collins & Quillian, 1972).

The semantic model, which is based on SL information processing of human brain cortex, is developed accordingly. Different areas of the brain cortex are involved in the processing and are connected in a hierarchical manner. Low-level information from sense organs is first processed in the primary information-processing regions of the brain cortex and is then transferred to high-level regions for further processing, such as abstracting, integrating, and interpreting. The detailed description and illustration of this hierarchical structure are summarized in Figure 1.

*Figure 1. Hierarchical structure. Low-level areas in the hierarchy generate specific information that increases speed and contain further details, whereas high-level areas form stable spatial invariance, change slowly, and show high-level semantic object expression (Adapted from Yao et al., 2015)*

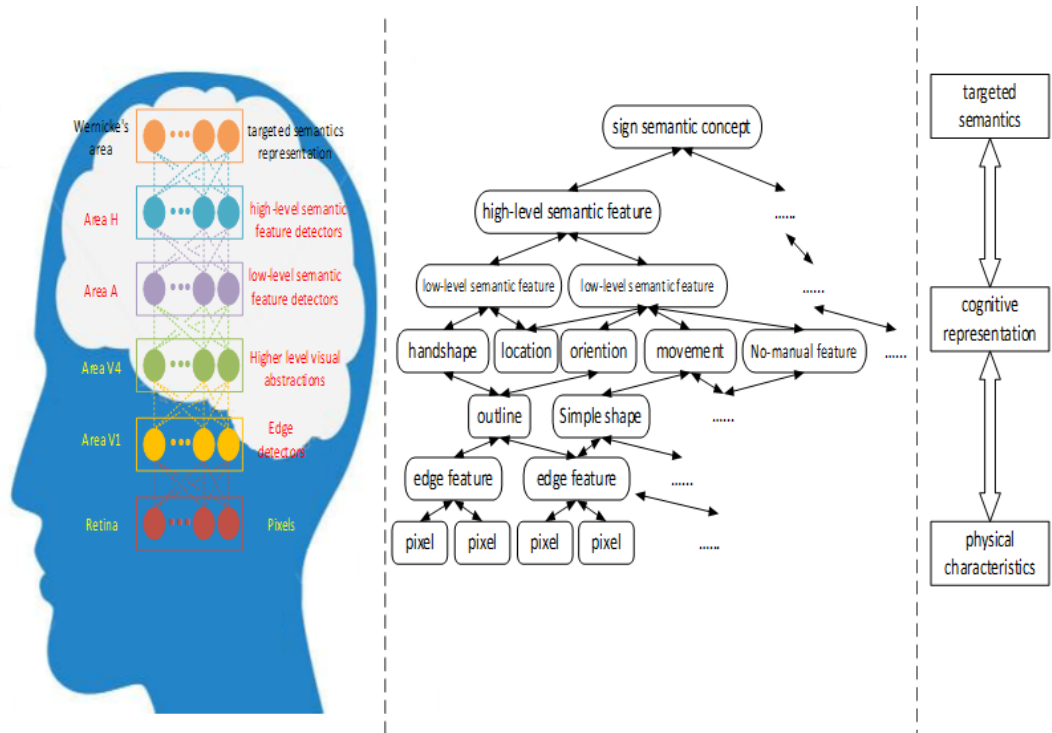

In SL communication, both substantial and semantic information (substantial information includes hand shape, hand location, hand movement, hand orientation, head tilting, shoulder tilting, eye gazing, body gestures, and facial expressions, and semantic information is represented into semantic concepts by these substantial SL information) almost exclusively relies on signs. However, accurate SL information analysis and prediction remain as challenging tasks in the field of natural SL processing. Three main tasks are, namely, capturing, decoding, and extracting the physical characteristics and relationship of signs (perception stage), matching the decoded cognitive representations with the stored semantic information (memory stage), and completing the machine translation process of SL information (judgment stage). This process of cognitive processing and understanding during SL communication is based on the PMJ principle of "from the definition and extraction/annotation of cognitive representation (Stage P) to the feature storage in line with the cognitive economy principles (Stage M), and then to the output of the classification and judgment (Stage J)."

The P→M→J (PMJ) principle exhibits a complete fine processing frame, the detailed illustration, and description of SL comprehension frame based on the PMJ principle is summarized in Figure 2.

*Figure 2. SL comprehension frame based on the PMJ principle. Perception refers to acquiring sign information through selective attention. The information is limitedly processed by the brain if prominence is given to useful and important information. Other information may be filtered out or suppressed when sources for information processing are limited. Memory refers to the spreading activation process, in which input information is coded, and one intends to store the information for a short period. Judgment*

*refers to the process in which the perceived information or the information stored in memory is compared, matched, or classified, and a decision or prediction is made. After the spreading activation, the network records the attention features of users and activates their future preferences (Yao et al., 2015)*

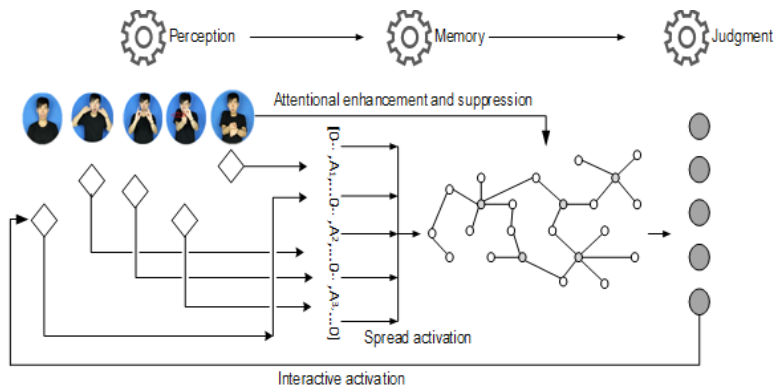

Concepts are in the form of network storage. The different concepts are stored in different functional areas in both hemispheres of the human brain. The same or similar concepts are stored in same or adjacent regions of brain. Specific information of entities in the outside world, such as humans, animals, or tools, is represented by the concept network in the human brain. This concept network (A concept is an abstract idea representing the fundamental characteristics of what it represents. Concept network consists of these abstract concepts) is, in turn, connected with the lexical network from mental lexicon in the left temporal lobe. Such specific information from mental lexicon will be employed to facilitate SL production during which the SL users generate classifier hand shapes under the guidance of the knowledge and rules of SL classifier predicates (Valli & Lucas, 2000). Here, classifier predicates are made by combining small meaningful unites to create bigger units, the main units being the hand shape and the movement. This condition implies that findings from brain research can provide knowledge and guidance for the cognitive computational modeling of classifier predicate comprehension. In order to obtain a deep understanding of sign lexical semantics, a cognitive processing model, which is based on the cognitive mechanism of human brain, is established. The cognitive processing model would activate the concept network of the associated classifier hand shapes in the brain. Here, classifier predicates differ from traditional linguistic units. Traditional methods, such as the syntactic tree, cannot satisfy the generation of the classifier predicates (Huenerfauth et al., 2006).

## DECISION-TREE BASED ALGORITHMIC METHODS

The authors use SNs as the knowledge representation and organization mode of SL. The relationship in semantic networks represents a type of information among nodes. Nodes with a complicated relationship with other nodes contain additional information. Such nodes require further effort to be understood. Consequently, the authors simulate selective attention (i.e., the processing of visual or auditory input based on whether it is relevant or important). They selected particular representations to enter perceptual awareness and therefore guide behavior. Through this process, less relevant information is suppressed by humans using the proposed algorithmic methods to accentuate the nodes selectively and suppress the unessential nodes (Chelazzi et al., 2013).

The emergence of 3-D-based sensors, such as *Kinect* by Microsoft and *Leap Motion* (Yao et al., 2014), has improved studies on sign recognition from video-based to 3-D-based sign recognition. However, this transformation makes traditional video-based SL recognition methods inapplicable to 3-D-based SL recognition technologies. Large training data are required for valid recognition in 3-D-based SL recognition technologies because of the low operation efficiency of the rotatable joint-based sorter and the matching techniques for sign signal recognition. Yao et al. (2014) proposed a decision tree-based algorithm. The algorithm aims to achieve a high-precision and real-time performance of SL automatic perception according to the features of *Leap Motion*. The authors adopted this method as the first step of SL comprehension.

## Attention Function

The authors propose the following attention function:

$$P_x = \frac{\sum_{all\ x} \Lambda_{ij}}{\sum \Lambda_{ij}} I_x \quad (1)$$

where $\sum_{all\ x} \Lambda_{ij}$ denotes the sum of the semantic relation weights around the semantic node $x$, $\sum \Lambda_{ij}$ denotes the sum of all semantic relation weights, and $I_x$ represents the activation value on the semantic node $x$ after the spreading activation process.

## Semantic Matching

Cognitive units in the memory network compete with one another based on certain rules to obtain more of the limited attention resources and more energy for a more active state. SL comprehension supports interactive activation models (Gutierrez et al., 2012). Therefore, judgment is the outcome of the attention competition game in the spreading activation, which is a search algorithm. The search algorithm is initiated by labeling a set of source nodes (e.g. concepts in a semantic network) with weights or "activation," and then iteratively propagating or "spreading" that activation out to other nodes linked to the source nodes processes of the human brain (Crestani, 1997; Preece, 1981).

A semantic matching algorithm based on activation spreading modes is proposed to determine the most appropriate semantic information. Activation starts to spread from the corresponding nodes of the signs presented by the signer. The activation value of the stimulus node (i.e., signs to be perceived before the start of spreading) must be calculated first. In particular, the increment in the interest value of object concept must be calculated, and this concept node must be used as an initial node for the spread study. Activation spreads to the neighboring nodes, which usually have a lower activation value than the source value. Therefore, introducing an activation attenuation factor for decreasing activation over the path length in the closed interval [0…1] is mandatory. That is, for every propagation through an edge a loss of activation is considered (Neumann et al., 1993; Rocha et al., 2004). The activation spreading process can be expressed as follows:

$$I_y(t+1) = O_x(t)\Lambda_{xy}(1-\delta)$$

$$(2)$$

where $I_y(t+1)$ represents that the value is spread from node $x$ to $y$ at time, $t+1$, $O_x(t)$ represents the activation value that was spread at node $x$ at time, $t$, $\Lambda_{xy}$ signifies the link between nodes $x$ and $y$, and $\delta$ is an attenuation factor used to describe the energy loss caused in the activation spreading process (Jiang & Tan, 2006).

Spreading activation theory states that the activation of human memory "chunks" (the content of any buffer is limited to a single declarative unit of knowledge, called a chunk) is determined by two factors (Anderson et. al., 2004; Anderson, 2013), namely, the use history of the memory chunk and the correlation between the memory chunk and the current retrieval information. These two factors calculate the weights and determine whether the chunk is activated and selected. This assumption has been verified by experimental cognitive psychology, and the calculation model has been established (Roelofs, 1992). The authors must use moments to express the distance in each activation time with the current time. Time units may be per hour as a unit and may also be the day. With the day as a unit, we can count the historical value in the previous day as the activation value of the first day. The algorithm based on the theory of memory activation can improve SL understanding, which is sometimes highly sensitive to time.

At node $y$, the largest number of neighbor nodes is ($n$-1); thus, the maximum of $I_y(t+1)$ can be expressed as:

$$I_y(t) = [I_1, I_2, \ldots, I_n]^T$$

i.e., the initial value of the semantic network. Where $I_1$, $I_2$, ..., $I_n$ are these activation value of neighbor nodes.

If activation spreads from a node in many directions, then its adjacent nodes obtain a low activation value. The adjacent nodes give a feedback value of their resonance energy (i.e., contributing structure with the lowest potential energy) to the co-adjacent nodes after they absorb the activation value. The following equation is therefore used:

$$I_z(t+1) = O_z(t) + \sum_{\text{all actived } x} O_x(t)\Lambda_{xz}(1\text{-}\delta) \qquad (3)$$

where $O_z(t)$ denotes the activation value of node $z$ at time $t$.

Given that the quantity of activated information is limited, the nodes that obtain less resonance information are equivalently inhibited and are less likely to be activated. The activation value distribution in the resonance process conforms to the human attention model.

## Attention Game Process

Cognitive units in memory network compete with one another by certain rules to increase the possibility of obtaining more human attention resources and more energy that will improve activity. This phenomenon is called a game process. The authors use game theory (Myerson, 1997), which is the study of mathematical models of conflict and cooperation between intelligent rational decision-makers and attempts to achieve the largest cognitive gains with the least energy possible, as a reference to simulate the attention enhancement and suppression processes that are selective attention processes. In other words, when visually searching for a non-spatial feature or a perceptual feature, selectively enhancing the sensitivity to that specific feature plays a role in directing attention. When people are told to look for motion, then motion will capture their attention, but attention is not captured by motion if they are told to look for color (Reynolds & Chelazzi, 2004). Activated results consistent with cognitive features can then be obtained. The authors assume that the game contains $n$ nodes. $s_i'$ and $s_i''$ are the two selectable strategies for node $i$, and they represent the acceptance and non-acceptance of the change in the attention function (i.e., $s_i', s_i'' \in S_i$). The corresponding gain can be represented by $u_i'$ and $u_i''$, and $u_i', u_i'' \in U_i$. $N$ nodes are assumed to reach an agreement before participating in the game to introduce the Nash equilibrium (i.e., each node only selects a specific strategy). The authors let $s^* = (s_1^*, \ldots, s_n^*)$ represent the agreement, where $s_i^*$ is the strategy of the node $i$ specified in the agreement. Nodes comply with this agreement only when the benefit from complying with the agreement is larger than that from not complying. This agreement constitutes Nash equilibrium if any node abides by this agreement. Thus, the Nash equilibrium is written as follows:

$$u_i(s_i^*, s_{-i}^*) \geq u_i(s_i, s_{-i}^*), \qquad \forall s_i \in S_i$$

(4)

where the combination of strategy $s^* = (s_1^*, \ldots, s_n^*)$ is a Nash equilibrium. Given that other nodes select $s_{-i}^* = (s_1^*, \ldots, s_{i-1}^*, s_{i+1}^*, \ldots, s_n^*)$, $s_i^*$ is the optimal strategy of each node $i$ (Myerson, 1997).

The attention game process determines whether the nodes need adjustment or need to be changed on the basis of the attention function. The activation energy distribution will reach a state consistent with the human attentive distribution after adjusting the activated value distribution. Nodes of the spread SNs have their own activation energy threshold values. The source node in the attention game process that represents a presented sign has the maximum activation value O in the present SNs. All equidistant nodes will participate in the game based on the attention function. The nodes with low activation energy (defined as the minimum energy required to start a chemical reaction) of a reaction is denoted by $E_a$ and given in units of kilojoules per mole (kJ/mol) or kilocalories per mole (kcal/mol)), threshold must be removed through a screening process to prevent them from participating in the enhancement and suppression processes of activating the most likely node. In the proposed screening, the authors ignore the nodes with a significantly low activation value to be activated in the enhancement process instead of lowering the possibility for other nodes to be activated.

The difference between the attentive readjustment in the present attention game process and the previous attentive allocation causes the instability in the overall cognitive structure of users to decrease knowledge credibility. Thus, a new cognitive structure must be determined at a cost as follows:

$$\text{Cost}(t, i, s_i, u_i, SN) = \sqrt{n^{-1} \sum_{i=1}^{n} (I_i(t+1) - O_i(t))^2} \quad (5)$$

where $I_i(t+1)$ denotes the activation value that is conveyed from one node at time $t+1$ to node $i$, and $O_i(t)$ denotes the activation value of node $i$ at time $t$. Therefore, the total cost is attributed to the change in the activation energy of all nodes in the SN. The goal of judgment is to achieve the overall optimal gain with a minimal computing cost. The gain function in the attention game process must then be determined. As the optimal strategy for node $i$, $s_i^*$ must minimize the distribution change that refers to the distribution change in the activation values of the overall network changed by the decision. The amount of spreading activation energy is fixed in the total process of activation spread in the SN; thus, the semantic node energy enhancement must be accompanied by reduced node energy. The attention parameters are affected by the overall distribution change in activation energy. The activation energy enhancement increases the impossibility of activating this node. Such activation is the ultimate purpose of each node (i.e., the node obtains the gain). Accordingly, the gain function is presented as follows:

$$\text{Gain}(t, i, s_i, u_i, SN) = \frac{\left(\sum_{j=1}^{num(all\ x)} I_{x \in \{neighbor\ node\}}(t+1) - \sum_{j=1}^{num(all\ x)} O_{x \in \{neighbor\ node\}}(t)\right)(1-\delta)}{num(all\ x)} \quad (6)$$

where SN represents the current semantic network, *num(all x)* represents the number of neighbor nodes $x$ of node $i$, $\sum_{j=1}^{num(all\ x)} O_{x \in \{neighbor\ node\}}(t)$ denotes the sum of the activation value that was spread of all node $i$ neighboring nodes at time $t$, the gain function is expressed as the attention gain of neighbor nodes $x$ of node $i$ after the enhancement and suppression processes, it represents the benefit a node gets by unilaterally changing their strategy.

The utility function of the attention game process can be determined as follows:

$$\text{Max}(u_i(t, s_i^*, s_{-i}^*)) = \text{Gain}(t, i, s_i, u_i, SNN) - \text{Cost}(t, i, s_i, u_i, SNN) \quad (7)$$

where $s_i^*$ is the optimal strategy of each node $i$, $s_{-i}^*$ is the strategies set of other nodes except node $i$. only when $u_i(t, s_i^*, s_{-i}^*)$ reaches the maximum, $s_i^*$ is a Nash equilibrium of node $i$. The utility of the other nodes will be affected by the decision of all other nodes because of the fact that the total quantity of activation energy is fixed (i.e., attention is limited) in the attention game process. When each node selects a decision for itself, it also considers the possible decision of other nodes and selects the "Nash equilibrium point" with maximum utility. This scenario is consistent with classical game theory. The authors select a Nash equilibrium decision for each node through the utility function of the attention game that is defined by Equation (7).

## METHODS

### Data Sets and Experimental Settings

All data from the authors' experiments are obtained from the Tsinghua University–Chinese SL Corpus (TH–SLC). The data mainly comprise SL expressions of idiom stories and life fragments of deaf students. No automatic annotation software based on videos is currently available because the annotation process for SL videos is time consuming and requires expert knowledge in dual language (i.e., Chinese language and Chinese SL). Video annotation is also time consuming. Specifically, it takes about 30 hours for the annotation RTF (real-time factor) of a parliamentary speech (i.e., One hour of speech requires 30 hours of annotation). However, the annotation RTF (real-time factor) for a full annotation of all manual and non-manual components of an SL video can reach up to 100 hours (Dreuw & Ney, 2008a). Therefore, such a corpus is significantly small. For example, the Aachen Boston database contains American SL and has annotated 201 English sentences (Dreuw & Ney, 2008a). The authors spent a year collecting more than 2000 sentences, but only 416 sentences containing 2496 signs were marked.

The authors asked 20 deaf students to select 300 sign pairs from 2469 annotated signs in TH–SLC and to judge the relevance of the sign pairs. The correlation values range from 0.0 to 1.0. For convenience, a five-point scale is used to assess the correlation. The sign pairs were obtained using a marked correlation. The authors establish an SN based on the word similarity computing method of *HowNet* (Liu & Li, 2002) to determine the connection weight of the network to validate the effects of the proposed model. The authors introduce the continuous bag-of-words (CBOW that predicts the current word from a window of surrounding context words. The order of context words does not influence the prediction (CBOW assumption) model (Mikolov et al., 2013), and the *HowNet* (Liu & Li, 2002) method as the baseline methods using the same recommended parameters. The efficiency of the utility function of the attention game process is evaluated in terms of word correlation computation, and the model complexity is analyzed.

### Word Relatedness Computation

Each model in this task needs to compute the semantic correlation of the given sign pair. The correlation between the experimental results of the model and human judgment reflects upon the model's performance. The authors selected 290 signs for the closed set and 10 signs for the open set.

Spearman's correlation between model correlation score and human judgment correlation score was calculated for comparison. Spearman correlation coefficient is defined as the Pearson correlation coefficient among the ranked variables (Myers & Well, 2003). For a sample of size *N*, original data $X_i, Y_i$ are converted into grade data $x_i, y_i$, the correlation coefficient $\rho$ is defined as follows:

$$\rho = 1 - \frac{6 \sum d_i^2}{n(n^2-1)}$$

(8)

where the difference between the observations of the two variable levels is set as $d_i = x_i - y_i$. If there is no duplicate value in the data, and two variables are completely monotonic correlation, the Spearman correlation coefficient is +1 or -1.

## RESULTS

For CBOW, the correlation scores of the two words are calculated using the cosine similarity of word embedding (Mikolov et al., 2013). The evaluative results of the baseline methods and the proposed SNM method in the closed test and in all test sets are shown in Table 1.

*Table 1. Evaluative results*

| Data Set | Closed Test | All Test Sets (Including Open Test) | Spearman's Rank Correlation Coefficient |
|---|---|---|---|
| Method | 290 pairs | 300 pairs | |
| CBOW (baseline method) | 0.4843 | 0.4869 | 0.4136 |
| Word similarity computing based on *HowNet* | 0.6157 | 0.6174 | 0.6052 |
| Proposed SNM method | 0.6951 | 0.7063 | 0.6437 |

The evaluation results show that the proposed SNM method is better than the baseline method in 290 and 300 word pairs. This finding indicates that the cognitive mechanism of sign comprehension is essential to understanding the meaning of signs. The internal structure, such as location, orientation, hand shape, and movement, contains rich semantic information. However, deep learning methods, such as CBOW, consider the external context, but ignore the internal structure.

Using the computing method of word similarity based on *HowNet* results in only a rough semantic computation. For example, adding 10 new sign pairs negligibly changes the performance of these methods. In other words, these methods can still handle new signs with improved performance. The semantic correlation of these new sign pairs calculated by the proposed method is close to human judgment. Figure 3 shows the quantitative analysis of the attention game process for two signs. Each hand shape of the two signs has at least 20 related semantic lexicons. The stimulus information and permutation of each node are shown in the first and second columns from high to low according to the activated value after the activation spreading process. Only 10 semantic lexicons that are maximally activated are shown. The permutation of each node is shown in columns three to seven from high to low according to the activation value after the end of the first to fifth attention games. The top 10 lexicons are also shown. The semantic lexicons in the blue background rank high after the games, those in the green background rank low after the games, and those in the white background are unchanged.

*Figure 3. Examples of attention games. The semantic lexicons in the blue background rank high after the games, those in the green background rank low after the games, and those in the white background are unchanged. This trend shows that the ranking of other semantic lexicons below slightly changes after the semantic lexicon that ranks highest becomes unchanged. This condition is due to the source that corresponds to the attention model being determined after several game processes.*

| input | activation value sorting after spread activation | activation value sorting after the first attention game | activation value sorting after the second attention game | activation value sorting after the 3rd attention game | activation value sorting after the 4th attention game | activation value sorting after the 5th attention game |
|---|---|---|---|---|---|---|
| 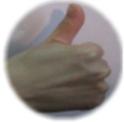 | A handshape Good Defend Maintain Protect Beheaded Support General Teacher Reliable | good Reliable Advanced Strange General Teacher Marshal Madam Ancestors Defend | General Beheaded Support Keep General Teacher Reliable Advanced Protect Maintain | General Beheaded Support Keep General Teacher Reliable Advanced Protect Maintain | General Beheaded Support Keep General Teacher Reliable Advanced Protect Maintain | General Beheaded Support Keep General Teacher Reliable Advanced Protect Maintain |
| 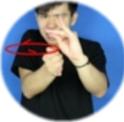 | Y handshape stand run lie Resistance Protest Control Incite Exploitation Recalcitrant | human animal burial Frustration future Voltage Ambassador coach Guide Blind | Animal Burial Frustration Future Voltage Ambassador Coach Guide Blind Opponent | cat dog horse human Ambassador coach stand run lie Resistance | cat dog horse human Ambassador coach stand run lie Resistance | cat dog horse human Ambassador coach stand run lie Resistance |

Figure 3 also shows that significant changes occur during the ranking of the semantic lexicons in the first and second instances after the first several games, whereas only a few changes occur in the following stimulus games. This trend shows that the ranking of lower semantic lexicons slightly change after the semantic lexicon that ranks highest becomes unchanged. This condition is due to the source that corresponds to the attention model being determined after several game processes. Attention is also assigned to other nodes in accordance with the attention game process. Humans reach a steady state after thinking about problems constantly, and the result negligibly changes if they rethink. Nearly no change is observed in the result after several rounds. Several semantic lexicons related to the signs are contained in the text set; thus, a few possible changes occur. The result of the attention game model conforms to human cognitive rules to a certain degree.

Attention is also assigned to other nodes in accordance with the attention game process (here, efforts have been made in modeling according to the mechanism of human attention). The result of the SNM conforms to human cognitive rules to a certain degree (Gutierrez et al., 2012). For example, the authors assume that deaf people understand the signs shown in Figure 3. Deaf people usually search for many familiar and specific nouns or signs in a spreading activation mode to comprehend classifier predicates. After all activated values are calculated; the activated nodes are graded and sorted. A high-activated value of the node indicates the importance of the interested object or concept represented by the node. This shows that deaf people are familiar with the concept node. Similar to the attention game process shown in Figure 3, the high-ranked semantic lexicon is a cat or dog after several rounds. This result shows that the most common subjects for deaf people are typical subjects that represent classifier predicates.

## DISCUSSION

Compared with that of existing models, the complexity of the proposed model is reflected mainly on the computational cost of the memory stage and the judgment stage (i.e., the computational cost of spreading activation and the attention game at time ($t$ + 1)). The cost is a dynamic value and related to two factors, namely, the activation state of the current sign and the current cycle as the first activation of the sign. Therefore, the value changes regardless of the choice of the user. This outcome is consistent with the strong dynamics of sign information, which can reflect the influence of information in different periods. In the memory stage, the time complexity of computing $O_x(t)$ is unity; thus, the time complexity is related to the total amount $N$ of activation energy and cycle times. The time complexity of each activation in each cycle is n × 1 = n. Space complexity is the storage space of each node and the semantic relation weight according to semantic similarity (semantic similarity can be estimated by defining a topological similarity, by using ontologies to define the distance between terms/concepts). Therefore, unlike the general model such as cobweb theorem model and vector space model, where the SNM increases the overhead in time complexity and space complexity. The model also increases the matching time of query nodes and weights in the current activation. However, the overhead at this time can provide more effective results than an invalid spreading and can be accepted by users.

In the judgment stage, when the node selects the game strategy to change its activated energy value, the convergence speed of adjusting the cognitive benefits to its own utility maximum "Nash equilibrium" is an important measure of evaluating the SNM (i.e., the cycle times of an attention game process). For the attention game, the Nash decision of different semantic nodes must minimize the change cost of the activation energy distribution of the entire network. The Nash equilibrium point decision for each node is selected using the utility function defined in the SNM. This process is repeated until the overall network activation energy distribution change is less than the specified threshold. The node needs to solve n-order nonlinear equations in every cycle. Therefore, the performance of the convergence speed of the SNM is indicated by the number of game cycles that the network requires to reach the Nash equilibrium point (i.e., the computing times of calculating the corresponding equation by each node in a game process). The square root of the sum of the variance of activation value $O_i(t + 1)$ of each adjusted node is directly reflected by the rate of convergence in the game process.

To verify its effectiveness, the attention game model is compared with the traditional model in terms of load balancing. In the traditional method, the activation value of each node is certain (i.e., the value is not enhanced or inhibited). The experimental results are shown in Figure 4. The results show that the load balance performance of the attention game model is better than that of the traditional model because the attention game model adjusts the activation strategy after the activation of each node. When the change cost of the energy distribution of the entire network activation is larger than the specified threshold, the human brain adjusts the strategy to inhibit the activation energy value in the next cycle. In doing so, the free competition and distribution of attention for each node according to the attention game model can be assured. The result is obtained through the overall competition. The load of attention of the network is balanced. The traditional model assumes that the activation energy value of each node is certain because the brain activation energy resource amount is constant in a period of time. The brain selects the node with a low activation energy value and performs the allocation of attention. This allocation causes the attention load of several nodes to be excessively large or unutilized.

*Figure 4. Comparison of load balance. The load balance performance of the SNM is better than that of the traditional model because the SNM adjusts the activation strategy after the activation of each node*

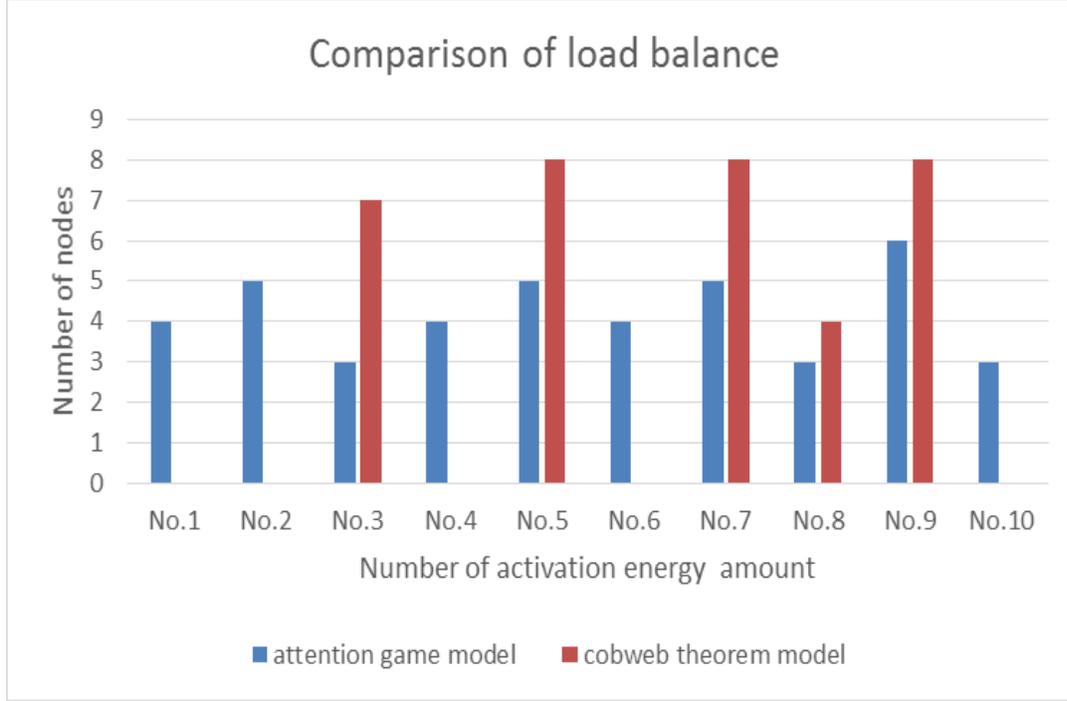

The proposed SNM model used Nash equilibrium to simulate the energy activation process. In order to quantitatively analyze the effects of Nash equilibrium, the authors compared the SNM with the cobweb theorem model (Pashigian, 2008) in terms of different activation energy amounts. The cobweb theorem is expressed as follows:

$$O(t + 1) = I(t) + r\left(D(O(t)) - S(O'(t))\right)$$

(9)

where $r$ is the adjustment parameter of the activation value, $D(O(t))$ is the activation function of a node, $O(t)$ is the activation value at time $t$, $S(O'(t))$ is the attention allocation function, $O'(t)$ is the expectation activation value at time $t$, and $D(O(t)) - S(O'(t))$ is the excessive demand function that represents the actual gaps between the activation value and activated allocated value. A large gap indicates a high activation value of the Nash Equilibrium of node. The parameter ($r$) indicates the actual speed and strength of adjusting the activation value according to the attention distribution condition in the last moment. When $r > 0$ it indicates that the adjustment direction of the activation value is consistent with the direction of the demand function.

The amount of activation energy $E_a$ is assumed to be 100 kJ/mol. Figure 5 shows the result of comparing the attention utilization between the game model and the cobweb model. The attention amount (attention is the behavioral and cognitive process of selectively concentrating on a discrete aspect of information, while ignoring other perceivable information. Attention amount refers to as the allocation size of limited processing resources), is less than 100 kJ/mol. If the attention amount is insufficient, then attention resources can only meet part of the node demand, and the resource utilization rate of the SNM will become higher than that of the cobweb model. When attention supply exceeds the demand of a node, the cobweb model achieves balance to meet the needs of several nodes after a repetitive cycle. The SNM meets the needs of all nodes, and the utilization rate of attention resources is higher than that of the cobweb model.

*Figure 5. Comparison of activation energy values. After the change in the initial value of the activation energy, the number of iterations increases depending on the difference between the initial activation energy value in the cobweb model and the balanced energy value. The iteration of the attention game model can be adjusted according to the difference in the activation energy between supply and demand. A sizeable adjustment is required to reach the Nash equilibrium state if a large difference exists between the supply and demand*

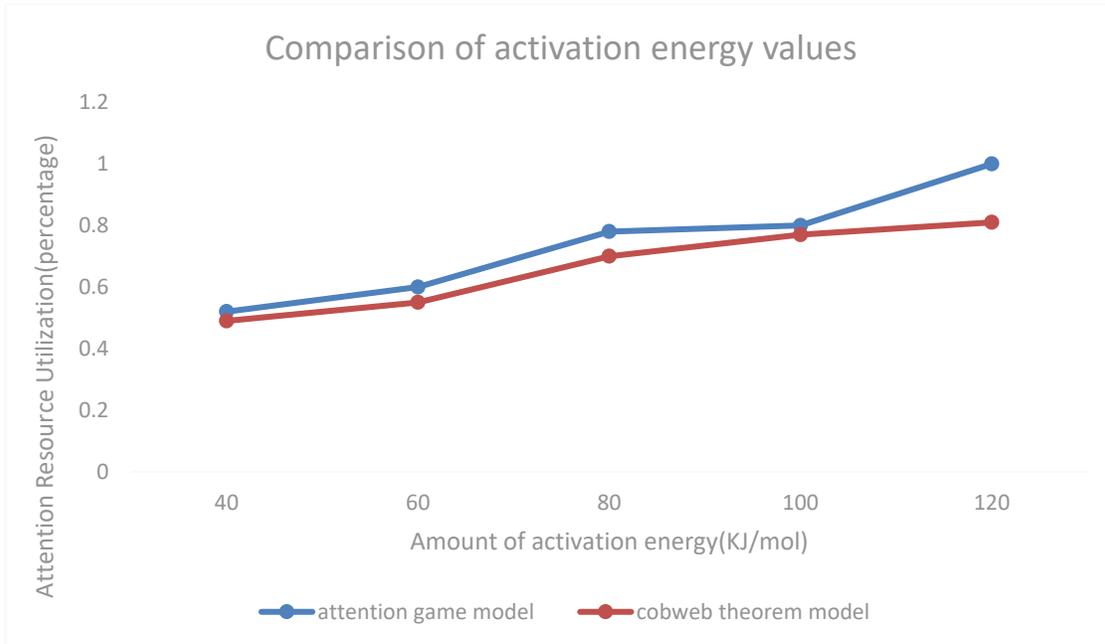

Figure 6 shows the cycle times of the SNM and the cobweb model that needs to achieve the Nash equilibrium. As shown in the figure, the equilibrium activation energy value of the nodes is 20 kJ/mol in the SNM, and the activation energy is 120 kJ/mol in total. If the initial value of the activation energy is changed, then the initial activation energy value of the cobweb model is higher than the energy equilibrium value and requires abundant cycle time. The SNM in each cycle can adjust the activation energy according to the variance of the activation energy. The variance and adjustment range are large, and the SNM eventually reaches the Nash equilibrium point.

*Figure 6. Cycle times of the SNM and the cobweb model that are needed to achieve the Nash equilibrium. When the supply falls short of demand, attention resources can only meet the demands of several nodes, and the resource utilization rate of the SNM becomes higher than that of the cobweb model. If the supply exceeds demand, then the cobweb model can reach equilibrium after repeated iterations and can meet only part of the demands of nodes. However, the SNM can meet the demands of all nodes, and its resource utilization rate is higher than that of the cobweb model*

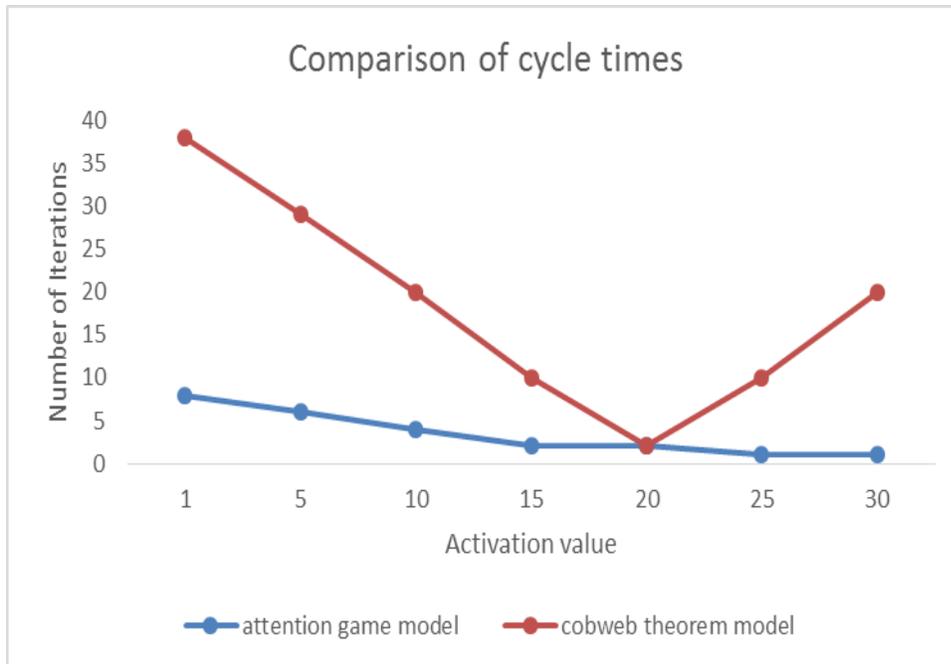

## CONCLUSION

The authors presented a new model for SL comprehension based on spatial information. This process uses game theory to simulate the human attention suppression and enhancement process. This process also joins the forgetting function of human memory traces to compute the initial state of the node. Memory is encoded with specific (semantic) meaning, or refers to information that is encoded along a spatial and temporal plane. Although the semantic network provides a functional view of how knowledge may be organized in the brain, it does not provide a clear model of how semantic memory might be presented in the brain (see Cacha et al., 2017). Spreading activation reveals that information can be stored in SNs for a long time, in which a network node is a linguistic concept and the nodes are connected through the correlation. An algorithmic method is proposed according to selective functions, and its effectiveness was verified using an example. The results show that the proposed method improves the performance of SL comprehension.

## ACKNOWLEDGMENT

The authors would like to thank Chunda Liu from the National Center for Sign Language and Braille for helping in stimulus preparation and data collection. This paper forms an expanded and revised version of a conference paper at the 14th IEEE International Conference on Cognitive Informatics & Cognitive Computing (ICCI* CC) at Tsinghua University, Beijing, July 6-8, 2015. The authors are grateful to Dr. Raymond Chiong, and two anonymous referees for their helpful comments.

### Conflict of Interest

The authors of this publication declare there is no conflict of interest.

### Funding Agency

This research was supported by the Beijing Municipal Natural Science Foundation [4202028]; National Social Science Foundation of China [21BYY106]; National Natural Science Foundation of China [62036001, 61866035, 61966033]; Premium Funding Project for Academic Human Resources